\documentclass{svproc}
\usepackage{url}

\usepackage{graphicx}
\usepackage{float}
\usepackage{booktabs}
\usepackage{siunitx}
\usepackage{threeparttable}
\usepackage{adjustbox}
\usepackage{placeins}

\usepackage[ruled,vlined]{algorithm2e}

\usepackage{xcolor}

\definecolor{idblue}{RGB}{33,113,181}
\definecolor{wrongred}{RGB}{203,24,29}
\definecolor{hallorange}{RGB}{255,127,0}
\definecolor{accgreen}{RGB}{35,139,69}

\usepackage[table]{xcolor}

\definecolor{idblue}{HTML}{1F4E79}      
\definecolor{wrongred}{HTML}{8B1E1E}    
\definecolor{hallorange}{HTML}{A65E00}  
\definecolor{accgreen}{HTML}{1E6B3A}    

\begin{document}
\mainmatter              
\title{Entropy and Attention Dynamics in Small Language Models: A Trace-Level Structural Analysis on the TruthfulQA Benchmark}
\titlerunning{Contribution Title}  
%
\author{Adeyemi Adeseye\inst{1}\and Aisvarya Adeseye\inst{2} \and Hannu Tenhunen\inst{3} \and Jouni Isoaho\inst{4}}
\authorrunning{A. Adeseye et al} 
%
%
\institute{Brilloconnetz Partners avoin yhtiö, Turku, Finland\\
\email{adeyemi@brilloconnetz.com}\\ 
\and
University of Turku, Turku, Finland\\
\email{aisvarya.a.adeseye@utu.fi}
\and
Royal Institute of Technology, Stockholm, Sweden\\
\email{hannu@kth.se}
\and
University of Turku, Turku, Finland\\
\email{jouni.isoaho@utu.fi} \\
}

\maketitle              

\begin{abstract}
                      |

Small language models (SLMs) have been increasingly deployed in edge devices and other resource-constrained settings. However, these models make confident mispredictions and produce unstable output, making them risky for factual and decision-critical tasks. Current evaluation methodology relies on final accuracy or hallucination rates without explaining how internal model behavior affects outputs. Specifically, how entropy evolves during decoding, how attention is distributed across layers, and how hidden representations contribute to uncertainty, logical inconsistencies, and misinformation propagation are often overlooked. Consequently, this study introduces a trace-level analysis of entropy and attention dynamics in SLMs evaluated with the TruthfulQA dataset. Four models with parameter ranges of 1B-1.7B parameters were examined via token-level output entropy, attention entropy, head dispersion, and hidden-state representation. The results reflect three model classifications by entropy patterns. Deterministic models (DeepSeek-1.5B and LLaMA-1B): output entropy decreases over time. Exploratory models (Gemma-1B): with increasing entropy, and balanced models (Qwen-1.7B): have moderate and stable entropy. Also, each group has distinctively different hidden-state movement and attention dispersion patterns. The analysis demonstrates that truthfulness in SLMs emerges from structured entropy and attention dynamics. Monitoring and optimizing these internal uncertainty patterns can guide the design of a more reliable, hallucination-aware, and application-specific edge SLMs.

\keywords{Small Language Models (SLMs), Entropy Dynamics, Attention Mechanisms, Truthfulness Evaluation, Uncertainty Analysis}
\end{abstract}
\section{Introduction}

Small language models (SLMs) are now often utilized in edge devices, privacy-sensitive domains, and other resource-constrained environments, requiring computational efficiency to coexist with reliability \cite{sharshar2025vlmedge,qu2025mobileedge}. In this study, SLMs refer to transformer-based autoregressive language models within the 1B–1.7B parameter size range \cite{wang2025slmsurvey}. Unlike large cloud-based systems, SLMs often operate in settings that need uncertainty controlled, hallucination minimized, and generation stability. However, language models can confidently produce incorrect outputs, propagate misinformation, and the generation of logically incoherent output \cite{adeseye2026hallucination,augenstein2024factuality,shah2025disinformation}. In safety and decision-critical environments, this leads to epistemic risk. Most evaluation frameworks focus on outcome-related metrics such as accuracy \cite{adeseye2025promptframework}, hallucination rate \cite{adeseye2026hallucination}, and calibration errors \cite{xia2025calibration}. These are important metrics. However, they do not explain how the internal model dynamic results lead to an output. Precisely, analysis that focuses on entropy evolution during decoding, attention dispersion across layers, and hidden-state representations remains limited \cite{skean2025layer}. These comparison at the token and layer levels enables comparison between different models for interpretation and systematic regulation \cite{apidianaki2023word}.

In transformer models, text generation relies on attention mechanisms and hidden representations \cite{vig2019attention}. While prior work has analyzed attention head redundancy and specialization \cite{voita2019multihead}, contextual representation geometry \cite{li2025geometry}, structural hidden state properties \cite{servedio2025hidden}, uncertainty in language models via entropy and calibration measures \cite{liu2025uncertainty}, these are done separately. There remains a lack of study that connects attention and representation dynamics to factual reliability \cite{wang2024monitor}. Entropy is usually treated as a scalar rather than a temporally evolving process \cite{zhu2026edis}. Consequently, the structure that governs generative truthfulness remains insufficiently understood, especially for SLMs, where architectural differences significantly affect model stability. 

Therefore, this study introduces a trace-level structural analysis of internal entropy and attention dynamics in SLMs on TruthfulQA benchmark. By trace-level structural analysis, we refer to the examination of the decoding steps' token-level output entropy, attention entropy, and hidden-state L2 magnitude rather than the final model correctness. To investigate how internal representation behavior maps to uncertainty and truthfulness outcomes, we analyze temporal entropy transition and cross-metric structural relationships. The study makes 3 major contributions. First, extension of static entropy measurement to a temporal, trace-level framework that captures the evolution of uncertainty during decoding. Second, probabilistic measures (entropy) are integrated with geometric measures (hidden-state movement), demonstrating the link between two partially coupled but structurally distinct dimensions. Third, we provide a generic classification of SLMs behavior (deterministic, exploratory, and balanced) based on internal uncertainty and stability characteristics. The linkage of truthfulness evaluation to structural properties advances a uniform framework to analyze SLMs' generative reliability, human evaluation, and reliability. This could guide future design of a more reliable and application-specific edge SLMs.


\section{Related works}

Truthfulness and hallucination research reveal today's language model tendency to generate fluent but factually false responses, especially with misleading or adversarial prompts \cite{brunello2026trustworthiness,ahmadi2024hallucination}. Also, most studies utilize contradiction detection and accuracy metrics in measuring final output corrections \cite{galitsky2024truthometer,chandler2024deep} without explaining the role of decoding dynamics when models fail confidently. This study performs and analyze token-level entropy and structural behavior to fill this gap. Numerous existing studies, such as uncertainty estimation, align the model's confidence with correctness \cite{shorinwa2025uncertainty,geng2024confidence}. In other studies, predictive uncertainty is measured by Shannon entropy \cite{karaca2022shannon,ray2021temperature}, while temperature scaling is explored to reduce overconfidence \cite{wen2024overconfidence,xu2025confidence}. Recent work has applied these techniques \cite{xie2024ats,kruse2025muse} with entropy treated as a single final value. However, none of this study measures how entropy changes during decoding steps or how these transitions result in factual reliability.



A wide range of studies have focused on attention mechanisms and interpretability \cite{rosser2026stream,ranaldi2025mechanistic}, some studies focused on attention heads redundancy and specializations \cite{ma2025cognitivemirrors,ling2025domainspecialization}, other studies investigated if attention weights explain model decisions \cite{chen2024xplainllm,artzy2024attend}. Attention layers also encode syntactic and semantic information \cite{faye2023cats}. However, there are limited studies that link attention entropy and output entropy together, which limits understanding of how attention patterns relate to prediction confidence, which this study fills. Other previous studies on representation geometry indicate that transformer layers have a structured but non-uniform transformation \cite{pavlovic2016transformers,lin2024normaltransformer}. Also, similarity analysis reveals layer specialization \cite{kharyuk2026specialization}, while mechanistic interpretability investigates functional components. Nevertheless, Uncertainty or truthfulness is rarely linked with geometric properties; they are usually separately studied. This work connects representational drift, hidden-state magnitude, and entropy evolution to better understand reliability.


Studies have focused on reasoning consistency in language models \cite{elazar2021consistency,braverman2020calibration}, but few measure how entropy or attention change over time. Our study models this across decoding steps to examine structural differences between the models. Finally, SLMs have been increasingly deployed in resource-constrained or privacy-sensitive domains \cite{alhindi2025ppsl,murala2025microservice}. However, reliability research majorly focuses on bigger models, while smaller models are less examined. Consequently, this study focuses on smaller models (1B–1.7B parameters). Generally, while previous work on truthfulness, entropy, attention, and representation geometry has been conducted independently on larger models, this work integrates them into a uniform framework linking internal structural dynamics to factual reliability in SLMs.

\section{Experimental Design and Trace-Level Structural Extraction}

\subsection{Dataset (TruthfulQA)}

This study evaluates model behavior using the TruthfulQA benchmark (contains 790 questions). This is a dataset designed to generate truthful and not just merely plausible responses. It is appropriate for this study because it spans multiple areas such as science, health, economics, history, law, and cultural topics, among other categories as well. It also provides differentiation between correct, misleading, and partially correct answers, supporting analysis beyond just binary classifications.
All models were prompted with the same set of questions with greedy decoding to minimize sampling variability. No fine-tuning or calibration was applied, ensuring that results reflect inherent model behavior under the same default settings. In general, it serves as an examiner of how structural dynamics differ between truthful and misleading responses.

\subsection{Model Selection and Computational Configuration}

This study evaluates four transformer-based SLMs within the 1B--1.7B parameter range: \textit{Llama-3.2-1B-Instruct}, \textit{DeepSeek-R1-Distill-Qwen-1.5B}, \textit{Gemma-3-1B-it}, and \textit{Qwen3-1.7B}. The selection was driven by model family, training, internal scaling behavior, and architectural differences amongst similar-sized models. The goal was not to compare model scale but internal structural differences that influence uncertainty and reliability. Execution was performed on a system with 64GB RAM and 8GB VRAM using FP16. Greedy decoding was intentionally selected instead of probabilistic sampling to simplify reproducibility. The deterministic rule removes stochastic variability that was introduced via temperature or nucleus sampling. Consequently, the entropy values reflect intrinsic model uncertainty rather than just randomness injected during sampling.

\subsection{Human Evaluation and Reliability}

To determine whether a response is truthful requires human evaluation. All the generated responses from the model were independently assessed by 2 independent researchers who reconciled the output with the TruthfulQA dataset. Both researchers labeled all responses independently first. Afterwards, disagreements were reconciled via consensus. The inter-rater reliablity using Cohen’s kappa ($\kappa$), and Krippendorff’s alpha ($\alpha$). Agreement was strong (Percent Agreement = 88\%, $\kappa = 0.81$, $\alpha = 0.79$), indicating substantial reliability under standard interpretation guidelines. This confirms the reproducibility and classification consistency of the process. The results provided the foundation needed to measure the internal behavior of the models.


\subsection{Algorithmic of Trace-Level Extraction}
This section contains the algorithm \ref{alg:trace_extraction} that explains how the model's internal behavior was extracted.  A more detailed explanation can be seen in the appendix.
 \begin{algorithm}[t]
 \scriptsize
\caption{Trace-Level Structural Extraction for Small Language Models (SLMs)}
\label{alg:trace_extraction}
\DontPrintSemicolon

\KwIn{Model list $\mathcal{M}$, prompt $P$, max steps $S_{\max}$, attention flag $A$}
\KwOut{Per-model trace CSV files and summary CSV}

Initialize $\texttt{summary\_data}[\text{"Prompt"}] \leftarrow P$\;
Set $\texttt{device} \leftarrow$ CUDA if available else CPU\;
Set $\texttt{dtype} \leftarrow$ FP16 if CUDA else FP32\;

\ForEach{$m \in \mathcal{M}$}{

    Load tokenizer $T_m$ and model $m$\;
    Set $m$ to evaluation mode and disable gradients\;

    Build stop token set $Stop_m$ using:\;
    tokenizer EOS IDs, config EOS IDs, generation EOS IDs,\;
    common stop tokens, and model-specific identifiers\;

    Build input IDs $X$ using chat template if available;\;
    otherwise tokenize raw prompt\;

    Initialize empty list $\texttt{trace\_rows}$\;

    \tcp{Prompt Phase (step = 0)}
    $\texttt{out} \leftarrow m(X,$ use\_cache=True, output\_hidden\_states=True, output\_attentions=$A)$\;
    Extract layer-wise metrics from $\texttt{out}$ and append to $\texttt{trace\_rows}$\;

    Initialize $\texttt{gen\_ids} \leftarrow X$\;
    Initialize $\texttt{past} \leftarrow \texttt{out.past\_key\_values}$\;
    Initialize $\texttt{step} \leftarrow 0$\;
    Initialize $\texttt{recent\_tokens} \leftarrow [\ ]$\;
    Initialize $\texttt{collected\_gen\_ids} \leftarrow [\ ]$\;

    \tcp{Generation Phase}
    \While{True}{

        $\texttt{step} \leftarrow \texttt{step} + 1$\;

        $\texttt{out} \leftarrow m(\texttt{gen\_ids}[-1],$
        $\texttt{past\_key\_values}=\texttt{past},$
        use\_cache=True, output\_hidden\_states=True, output\_attentions=$A)$\;

        $x_{t+1} \leftarrow \arg\max(\texttt{out.logits}_{t})$\;

        Append $x_{t+1}$ to $\texttt{gen\_ids}$ and $\texttt{collected\_gen\_ids}$\;
        $\texttt{past} \leftarrow \texttt{out.past\_key\_values}$\;

        Extract layer-wise metrics from $\texttt{out}$ and append to $\texttt{trace\_rows}$\;

        Update $\texttt{recent\_tokens}$ (retain last 15 tokens)\;

        \If{15 consecutive tokens are identical}{
            \textbf{break}\;
        }

        \If{$x_{t+1} \in Stop_m$ \textbf{or} $\texttt{step} \ge S_{\max}$}{
            \textbf{break}\;
        }
    }

    Decode final output from $\texttt{collected\_gen\_ids}$\;
    Store in $\texttt{summary\_data}[m]$\;

    Save $\texttt{trace\_rows}$ to per-model trace CSV\;
    Free model and tokenizer memory; clear cache\;
}

Save $\texttt{summary\_data}$ to summary CSV\;

\end{algorithm}

\FloatBarrier

\section{Entropy-Based Structural Evaluation and Generation Dynamics}

\subsection{Cross-Model Structural Entropy and Attention Analysis}
Table \ref{tab:transposed_model_entropy} presents clear structural differences across the four SLMs during the TruthfulQA generation. The results show distinctively different entropy states, attention behavior, and decoding confidence. For generated token length (Gen), Qwen (431 tokens) and LLaMA (360 tokens) produce substantially longer outputs than Gemma (133 tokens) on average, while DeepSeek generated the least (26 tokens). The implication is that longer output increases structural trace depth. However, it is important to note that KV memory is not purely driven by generation length but affected by architectural factors such as hidden size and number of attention heads. This explains why Qwen showed an extremely large KV footprint despite not having an average lesser token than LLaMA and slightly more than Gemma and DeepSeek.

Shannon entropy reflects distributional spread (Shannon, 1948). Output entropy is a representation of Shannon entropy. DeepSeek produced highly peaked probability distributions, while Gemma distributed probability more evenly across candidate tokens. Gemma had the highest mean output entropy (0.678), while DeepSeek had the lowest (0.124). A more accurate output is usually associated with lower output entropy, while an incorrect answer is usually associated with higher entropy. DeepSeek was the most highly confident model (strong internal certainty) because of its low entropy, which aligns with a very high Top1 probability (0.970) and a large Top1–Top2 gap (0.957). Gemma and Qwen were moderately confident, while LLaMA sits in between. Also, output entropy distribution analysis indicates that for LLaMA and Gemma, the mean exceeds the median, which means most tokens had low entropy but occasionally exhibit very high uncertainty. Deepseek had the tightest distribution, where most of its tokens are almost deterministic. Qwen falls in the middle of both of these categories. Output entropy standard deviation further reflects model stability. Gemma (0.920) had the highest volatility, which suggests an exploratory decoding behavior, then LLaMA (0.722) and Qwen (0.527). DeepSeek (0.483) was the most stable

For attention entropy, Gemma had the highest mean attention entropy (2.402), LLaMA was moderate, while DeepSeek was the lowest. Usually, high attention points to diffused contextual focus, while lower attention denotes a more concentrated focus. Also, attention weights define how contextual information is integrated. Consequently, diffuse attention means support for broader evidence aggregation, whereas focused attention may reflect stronger token-level selectivity. This means Gemma generally had a broader scanning strategy while Deepseek generally had a narrower mean attention focus. Standard deviation measures stability during decoding steps. DeepSeek had the lowest SD (0.139), indicating highly stable attention behavior. Gemma (0.255) is moderately stable, followed by LLaMA (0.292), which exhibited slightly more fluctuation, while Qwen had the highest SD (0.325). This means DeepSeek’s attention structure did not change very much across steps, while Qwen’s had the most dynamic fluctuations. Exhibiting stable attention suggests consistent structural allocation of contextual importance, whereas higher variability may indicate adaptive context shifting. The relationship between mean and median further clarifies distributional shape. For all models, mean $\approx$ median, which indicates that attention entropy distributions are roughly symmetric and stable, unlike output entropy, which exhibited strong right skew. Therefore, attention behavior did not exhibit extreme spikes in the same way that token-level output entropy uncertainty did.

According to multi-head attention theory, higher dispersion may indicate stronger head specialization and representational diversity \cite{huang2019multihead}. Head Dispersion Index (HDI) reveals significant architectural variation \cite{bakhtyari2022dispersion}. Qwen (0.951) had the highest dispersion, pointing to a more uniform attention heads behavior. LLaMA was also considerably high. Contrariwise, DeepSeek and Gemma had lower HDI values, which reflect a more uniform head behavior.

\begin{table*}[htbp]
\centering
\caption{Overall Generation and Entropy Summary (GEN Phase). 
Values are reported as Mean (SD). Entropy measures represent 
token-level Shannon entropy and are non-negative by definition.}
\label{tab:transposed_model_entropy}

\begin{adjustbox}{max width=\textwidth}
\begin{threeparttable}

\begin{tabular}{
l
S[table-format=3.3(3)]
S[table-format=3.3(3)]
S[table-format=3.3(3)]
S[table-format=3.3(3)]
}

\toprule
\textbf{Metric}
& {\textbf{LLaMA-3.2-1B}}
& {\textbf{Gemma-3-1B}}
& {\textbf{DeepSeek-1.5B}}
& {\textbf{Qwen-1.7B}} \\
\midrule

Prompt Tokens
& 46 & 19 & 16 & 23 \\

Gen Tokens
& 360 & 133 & 26 & 431 \\

Output Entropy
& 0.570(722)
& 0.678(920)
& 0.124(483)
& 0.374(527) \\

Top1 Prob
& 0.817(225)
& 0.806(220)
& 0.970(111)
& 0.875(182) \\

Top1--Top2 Gap
& 0.713
& 0.697
& 0.957
& 0.795 \\

Attn Entropy Mean
& 1.809(292)
& 2.402(255)
& 1.896(139)
& 1.966(325) \\

Head Dispersion (HDI)
& 0.871(128)
& 0.517(085)
& 0.495(078)
& 0.951(131) \\

KV Total MB (Max)
& 25.375
& 7.719
& 2.297
& 99.312 \\

\bottomrule
\end{tabular}

\begin{tablenotes}
\scriptsize
\item NOTE: Values are reported as Mean (SD), where the number in parentheses 
denotes the standard deviation across generated tokens. 
Output Entropy and Attention Entropy represent token-level Shannon entropy 
($H = -\sum p(x)\log p(x)$) and are non-negative by definition. 
Top1 Prob refers to the mean probability assigned to the selected next token. 
Top1--Top2 Gap denotes the average difference between the highest and 
second-highest token probabilities. HDI represents the Head Dispersion Index. 
KV Total MB indicates the maximum key-value cache memory usage during generation.
\end{tablenotes}

\end{threeparttable}
\end{adjustbox}
\end{table*}
\FloatBarrier

Table \ref{tab:transposed_stats} presents the other output entropy distribution profile not present in Table  \ref{tab:transposed_model_entropy}.




All models had P10 = 0 and Min = 0, indicating that at least 10\% of tokens were fully deterministic. All models exhibited these deterministic predictions during token generation. However, P90 values showed clear divergence.  LLaMA (1.571) and Gemma (1.472) frequently entered high-uncertainty states, with Qwen (1.079) being moderate and DeepSeek (0.018) being extremely low. This means 90\% of DeepSeek tokens are almost deterministic, whereas LLaMA and Gemma occasionally had strong internal competition between candidate tokens. Usually, high P90 values indicate unstable or exploratory moments in the decoding trajectory.



Maximum entropy helps capture rare extreme events. Gemma had the highest maximum (5.662), followed by LLaMA (3.706) and Qwen (2.947). DeepSeek had the lowest (2.376). This means Gemma has the highest uncertainty tail; it occasionally produces extremely uncertain tokens.  In factual benchmarks like TruthfulQA, such spikes may correspond to epistemic confusion or knowledge boundary cases.


Generally, three behavioural insights emerged from the distributional structure. DeepSeek operates in a highly deterministic manner, characterized by very low median entropy, very low P90, and a tight distribution. Gemma operates in a volatile or exploratory manner, with the highest mean entropy, highest SD, and strongest extreme spikes. LLaMA and Qwen operate in a balanced manner, with moderate entropy levels and moderate volatility.

From a theoretical point of view, entropy measures uncertainty in probabilistic prediction \cite{kang2016nonprob}. Consequently, very low entropy is a reflection of rigid confidence, which may increase the risk of confident errors. Also, very high entropy reflects unstable reasoning and token competition. Balanced entropy with controlled variance may support better calibration and structured reasoning during uncertainty. These distributional patterns provide structural insight into how SLMs manage epistemic uncertainty during TruthfulQA generation.

\begin{table*}[htbp]
\centering
\caption{Other Output Entropy Distribution Profile}
\label{tab:transposed_stats}

\begin{adjustbox}{max width=\textwidth}
\begin{threeparttable}

\begin{tabular}{
l
S[table-format=1.3]
S[table-format=1.3]
S[table-format=1.3]
S[table-format=1.3]
}

\toprule
\textbf{Statistic}
& {\textbf{LLaMA-1B}}
& {\textbf{Gemma-1B}}
& {\textbf{DeepSeek-1.5B}}
& {\textbf{Qwen-1.7B}} \\
\midrule




P10
& 0.000
& 0.000
& 0.000
& 0.000 \\

P90
& 1.571
& 1.472
& 0.018
& 1.079 \\

Min
& 0.000
& 0.000
& 0.000
& 0.000 \\

Max
& 3.706
& 5.662
& 2.376
& 2.947 \\

\bottomrule

\end{tabular}
\end{threeparttable}
\end{adjustbox}
\end{table*}
\FloatBarrier

Table \ref{tab:transposed_attn_stats} presents the other parameters from the distribution of step-wise attention entropy across models not found in Table \ref{tab:transposed_model_entropy}.

Percentile ranges confirm these structural patterns. Qwen shows a relatively wide P10–P90 range (1.475 → 2.347), indicating noticeable variability in attention dispersion across steps. DeepSeek shows a very tight range (1.759 → 1.998). This confirms that DeepSeek maintains a highly consistent attention structure throughout generation. Gemma maintains high entropy even at P10 (2.028), meaning even its lowest attention states remain relatively diffuse.

Minimum and maximum values reinforce these trends. Gemma reaches the highest maximum (2.995), suggesting occasional very broad context integration. DeepSeek operates within the narrowest band. From a theoretical standpoint, attention entropy captures distributional smoothness of attention weights. Controlled and stable attention may reduce structural noise, while excessive dispersion may dilute representational strength (Michel et al., 2019).

Overall, three attention regimes emerge. Gemma operates in a high-dispersion regime with consistently diffuse attention. DeepSeek operates in a concentrated and highly stable regime. LLaMA and Qwen sit in a moderate regime, with Qwen showing more fluctuation across steps. These findings indicate that attention dynamics are architecturally structured and differ independently from output entropy behavior. Such structural attention differences may influence how models integrate contextual evidence when responding to TruthfulQA prompts.

\begin{table*}[htbp]
\centering
\caption{Step-Wise Attention Entropy Distribution}
\label{tab:transposed_attn_stats}

\begin{adjustbox}{max width=\textwidth}
\begin{threeparttable}

\begin{tabular}{
l
S[table-format=1.3]
S[table-format=1.3]
S[table-format=1.3]
S[table-format=1.3]
}

\toprule
\textbf{Statistic}
& {\textbf{LLaMA-1B}}
& {\textbf{Gemma-1B}}
& {\textbf{DeepSeek-1.5B}}
& {\textbf{Qwen-1.7B}} \\
\midrule

P10
& 1.387
& 2.028
& 1.759
& 1.475 \\

P90
& 2.173
& 2.708
& 1.998
& 2.347 \\

Min
& 0.994
& 1.712
& 1.647
& 1.094 \\

Max
& 2.475
& 2.995
& 2.335
& 2.719 \\

\bottomrule

\end{tabular}
\end{threeparttable}
\end{adjustbox}
\end{table*}

\FloatBarrier

\subsubsection{Layer-Level Attention Structure}

Table \ref{tab:transposed_layer_entropy} presents the layer-wise entropy profile across models. This analysis reveals how attention dispersion behaves vertically across transformer depth.

The number of layers differs across architectures. LLaMA has 16 layers, while Gemma has 26, and DeepSeek and Qwen each have 28. This matters because deeper models can distribute representational roles differently across layers. Greater depth allows hierarchical abstraction and progressive refinement of information. In transformer theory, early layers often capture local relations, while deeper layers encode higher-level semantics \cite{kang2016nonprob}.

Mean layer entropy follows the global pattern observed earlier. Gemma maintains the highest average layer entropy (2.402), indicating consistently diffuse attention across depth. LLaMA has the lowest mean (1.809), suggesting more concentrated average attention. DeepSeek and Qwen sit between these extremes.

Structural imbalance across layers is captured by the standard deviation across layers. Qwen shows the highest SD (0.666), followed by LLaMA (0.600), Gemma (0.537), and DeepSeek (0.351). This shows that Qwen and LLaMA have stronger layer-wise variation, while DeepSeek is structurally more uniform across depth. Higher cross-layer variance suggests that different layers perform sharply distinct functions. Lower variance suggests smoother vertical transitions.

The gap between lowest and highest entropy layers highlights specialization. Qwen’s lowest entropy layer is 0.836, while its highest is 3.153. This is a very large internal range. It indicates strong specialization — some layers are highly focused, while others are highly diffuse. In contrast, DeepSeek’s range is much tighter. This confirms that Qwen exhibits strong vertical heterogeneity, while DeepSeek maintains consistent structural behavior across depth.

Mean HDI further reinforces this pattern. Qwen shows the highest HDI (0.951), followed by LLaMA (0.871). Gemma (0.517) and DeepSeek (0.495) are substantially lower. This means Qwen has the most uneven head behavior within layers, suggesting stronger head specialization. DeepSeek and Gemma show more uniform head behavior. According to multi-head attention theory, specialization across heads supports representational diversity \cite{huang2019multihead}. However, excessive heterogeneity may also increase structural instability.

When combining step-level and layer-level findings, two structural dimensions become visible. At the step level, Gemma is the most diffuse overall, while DeepSeek is the most stable across decoding steps. At the layer level, Qwen is the most structurally heterogeneous, while DeepSeek is the most structurally uniform. This indicates that architectural depth and head dispersion introduce a second axis of structural differentiation beyond token-level entropy.

From a theoretical perspective, transformers construct hierarchical internal representations across depth \cite{ahmadi2024hallucination}. Layer-wise entropy variation reflects how attention shifts from broad contextual encoding to refined information compression. Uniform vertical structure, as seen in DeepSeek, suggests controlled abstraction. Strong vertical heterogeneity, as seen in Qwen, suggests dynamic specialization across layers. These structural properties provide insight into how architectural depth shapes internal reasoning organization during TruthfulQA generation.

\begin{table*}[htbp]
\centering
\caption{Layer-Wise Entropy Profile}
\label{tab:transposed_layer_entropy}

\begin{adjustbox}{max width=\textwidth}
\begin{threeparttable}

\begin{tabular}{
l
S[table-format=2.0]
S[table-format=2.3]
S[table-format=2.3]
S[table-format=2.3]
}

\toprule
\textbf{Metric}
& {\textbf{LLaMA-1B}}
& {\textbf{Gemma-1B}}
& {\textbf{DeepSeek-1.5B}}
& {\textbf{Qwen-1.7B}} \\
\midrule

\#Layers
& \multicolumn{1}{c}{16}
& \multicolumn{1}{c}{26}
& \multicolumn{1}{c}{28}
& \multicolumn{1}{c}{28} \\

Mean Layer Entropy
& 1.809
& 2.402
& 1.896
& 1.966 \\

SD Across Layers
& 0.600
& 0.537
& 0.351
& 0.666 \\

Lowest Entropy Layer
& \multicolumn{1}{c}{L3 (1.015)}
& \multicolumn{1}{c}{L18 (1.313)}
& \multicolumn{1}{c}{L1 (1.196)}
& \multicolumn{1}{c}{L25 (0.836)} \\

Highest Entropy Layer
& \multicolumn{1}{c}{L1 (2.787)}
& \multicolumn{1}{c}{L9 (3.048)}
& \multicolumn{1}{c}{L8 (2.447)}
& \multicolumn{1}{c}{L2 (3.153)} \\

Mean HDI
& 0.871
& 0.517
& 0.495
& 0.951 \\

\bottomrule
\end{tabular}

\end{threeparttable}
\end{adjustbox}
\end{table*}
\FloatBarrier

\subsection{Temporal Entropy Dynamics and Representational Evolution}

\subsubsection{Early vs Late Generation Entropy Drift}

Table \ref{tab:transposed_early_late} examines how entropy changes from the first 20\% to the last 20\% of generation. The results show clear temporal transitions in both attention and output uncertainty.

Attention entropy increases for all models. LLaMA (+0.666), Gemma (+0.568), DeepSeek (+0.284), and Qwen (+0.715) all show positive shifts. This means attention becomes more diffuse toward the end of generation for every model. As generation progresses, attention spreads out more. This is a universal trend across architectures. In autoregressive transformers, later tokens must integrate longer context windows (Vaswani et al., 2017). Broader attention may reflect increased contextual aggregation.

Output entropy, however, splits into two distinct behaviours. LLaMA (-0.388) and DeepSeek (-0.526) show decreasing output entropy. This means these models become more confident over time. Their probability distributions sharpen, and they “lock in” toward the end of generation. DeepSeek shows the most extreme case. Early output entropy is 0.528, while late entropy drops to 0.002. This is almost deterministic. It suggests DeepSeek strongly commits to final tokens.

In contrast, Gemma (+0.489) and Qwen (+0.255) show increasing output entropy. This means these models become more uncertain later in generation. They shift into a more exploratory regime rather than consolidating confidence. This pattern may relate to factual instability in longer responses. As sequence length increases, uncertainty may accumulate if internal representations are not tightly stabilized.

Importantly, attention entropy increases for all models, but output entropy does not follow the same direction. This means internal attention diffusion does not directly determine prediction confidence. Attention dispersion and token-level certainty are related but not identical processes. Attention governs information integration, while output entropy reflects final probability allocation over vocabulary space.

From an information-theoretic perspective, entropy drift reflects dynamic changes in epistemic uncertainty (Shannon, 1948). Decreasing entropy suggests convergence toward a stable belief state. Increasing entropy suggests growing ambiguity or representational diffusion.

Overall, entropy is not static during generation. Models transition into different uncertainty regimes. Some models consolidate confidence over time (DeepSeek, LLaMA). Others amplify uncertainty (Gemma, Qwen). This provides strong evidence that generative behaviour differs dynamically across architectures. These temporal differences highlight that architectural design influences not only static entropy levels but also how uncertainty evolves during TruthfulQA generation.

\begin{table*}[htbp]
\centering
\caption{Early vs Late Entropy Shift (First 20\% vs Last 20\%)}
\label{tab:transposed_early_late}

\begin{adjustbox}{max width=\textwidth}
\begin{threeparttable}

\begin{tabular}{
l
S[table-format=1.3]
S[table-format=1.3]
S[table-format=+1.3]
S[table-format=1.3]
S[table-format=1.3]
S[table-format=+1.3]
}

\toprule
\textbf{Metric}
& {\textbf{LLaMA-1B}}
& {\textbf{Gemma-1B}}
& {\textbf{DeepSeek-1.5B}}
& {\textbf{Qwen-1.7B}} \\
\midrule

Output Entropy (Early)
& 0.865
& 0.508
& 0.528
& 0.245 \\

Output Entropy (Late)
& 0.477
& 0.996
& 0.002
& 0.500 \\

$\Delta$ Output
& -0.388
& +0.489
& -0.526
& +0.255 \\

Attn Entropy (Early)
& 1.453
& 2.053
& 1.766
& 1.533 \\

Attn Entropy (Late)
& 2.119
& 2.621
& 2.050
& 2.248 \\

$\Delta$ Attn
& +0.666
& +0.568
& +0.284
& +0.715 \\

\bottomrule

\end{tabular}
\end{threeparttable}
\end{adjustbox}
\end{table*}

\FloatBarrier
\subsubsection{Extremal Entropy Layers}

Table \ref{tab:low_high_entropy_layers} compares the lowest and highest entropy layers in LLaMA-1B together with their corresponding HDI values. The gap between the lowest entropy layers (1.015–1.378) and the highest entropy layers (2.321–2.787) is large, approximately 1.7. This is a substantial difference. It shows that some layers are strongly focused while others are highly diffuse. LLaMA does not treat all layers equally.

Early layers tend to be more diffuse. Layer 1 has the highest entropy (2.787), and several early–mid layers (7–10) also appear among the highest entropy group. This suggests that early and middle layers distribute attention broadly. They likely function as contextual information gatherers. In transformer theory, lower layers often encode broad lexical and positional relationships before deeper abstraction occurs \cite{zhang2025semantic}.

Focused layers appear in mid and deeper positions. Layers 3 and 2, as well as deeper layers such as 14, 13, and 16, show low entropy values. This suggests that some deeper layers become more concentrated. These layers may refine representations and consolidate information before final prediction. Such refinement behavior aligns with hierarchical representation theory in deep transformers.

HDI patterns further reveal head-level structure. The highest entropy layer (Layer 1) also has a very high HDI (1.099). This means that in diffuse layers, attention heads behave unevenly. Some heads dominate or specialize differently from others. In contrast, lower entropy layers show lower HDI values (0.708–0.922). This suggests that focused layers tend to have more uniform head behavior.

Together, these results demonstrate structural specialization. Entropy is not evenly distributed across layers. Some layers act as “broad context collectors,” while others function as “refinement or decision layers.” This reflects architectural specialization within LLaMA. According to multi-head attention theory, such differentiation supports representational diversity and hierarchical abstraction (Michel et al., 2019). The layer-wise entropy structure therefore provides clear evidence of vertical functional differentiation during generation.

\begin{table*}[htbp]
\centering
\caption{Comparison of Lowest and Highest Entropy Layers with Corresponding HDI for LLaMA-1B}
\label{tab:low_high_entropy_layers}

\begin{tabular}{
S[table-format=2.0]
S[table-format=1.3]
S[table-format=1.3]
S[table-format=2.0]
S[table-format=1.3]
S[table-format=1.3]
}

\toprule
\multicolumn{3}{c}{\textbf{Lowest Entropy Layers}} 
& \multicolumn{3}{c}{\textbf{Highest Entropy Layers}} \\
\cmidrule(lr){1-3} \cmidrule(lr){4-6}

\textbf{Layer} & \textbf{Entropy} & \textbf{HDI}
& \textbf{Layer} & \textbf{Entropy} & \textbf{HDI} \\
\midrule

3  & 1.015 & 0.821 
& 1  & 2.787 & 1.099 \\

2  & 1.162 & 0.922 
& 8  & 2.741 & 0.932 \\

14 & 1.171 & 0.766 
& 9  & 2.638 & 0.882 \\

13 & 1.352 & 0.708 
& 7  & 2.380 & 0.881 \\

16 & 1.378 & 0.728 
& 10 & 2.321 & 0.989 \\

\bottomrule
\end{tabular}
\end{table*}
\FloatBarrier

\subsubsection{Representation Magnitude and Transformation}

Table \ref{tab:hidden_l2_stats} reports the mean L2 norm of hidden representations and the average step-to-step change (Delta L2). These metrics describe the geometric magnitude of internal states and how strongly they transform during autoregressive decoding.

The scale differences are extremely large. Gemma-1B shows a mean hidden L2 of 6419.709, Qwen-1.7B shows 567.026, DeepSeek-1.5B shows 137.372, and LLaMA-1B shows 13.132. These values are not just different — they are on completely different scales. This indicates that hidden state magnitude is architecture-dependent and cannot be directly compared in absolute terms across models. Such differences reflect hidden dimension size, normalization strategy (e.g., RMSNorm vs LayerNorm), and internal scaling choices in model design. The L2 norm therefore captures representational energy rather than confidence.

Representation drift (Delta L2) shows how much hidden states move between consecutive tokens. Gemma again shows the largest drift (1600.649), followed by Qwen (303.196), DeepSeek (67.798), and LLaMA (9.735). This means Gemma transforms its hidden state much more aggressively between tokens, while LLaMA changes its representation more gently and smoothly. Larger drift suggests stronger internal reconfiguration per decoding step.

When connected to earlier entropy results, an important pattern emerges. Gemma, which showed higher output entropy and higher attention entropy, also exhibits the largest representational drift. This suggests that higher uncertainty is associated with larger internal state transformations. In contrast, DeepSeek, which showed very low output entropy and strong late-stage confidence consolidation, has much smaller drift compared to Gemma. This indicates tighter representational control.

However, representation magnitude and entropy are not identical constructs. Large hidden L2 does not automatically imply high entropy. Drift magnitude reflects transformation strength, not uncertainty directly. Internal geometric energy and probabilistic uncertainty operate as partially independent dimensions. Entropy measures distributional uncertainty over tokens, while L2 norms reflect vector magnitude in hidden space. One captures probabilistic dispersion; the other captures geometric dynamics.

Overall, the models operate in distinct representational regimes. LLaMA shows low-magnitude, low-drift dynamics with stable evolution. DeepSeek shows moderate magnitude and controlled drift. Qwen shows stronger movement. Gemma operates in a high-magnitude, high-drift regime. These geometric differences reinforce that generative behavior is shaped not only by entropy and attention structure but also by the internal geometry of hidden state evolution.

\begin{table}[htbp]
\centering
\caption{Hidden Representation Magnitude and Drift}
\label{tab:hidden_l2_stats}

\begin{tabular}{
l
S[table-format=4.3(3)]
S[table-format=4.3(3)]
}

\toprule
\textbf{Model}
& {\textbf{Hidden L2 (Mean $\pm$ SD)}}
& {\textbf{Delta L2 from Prev (Mean $\pm$ SD)}} \\
\midrule

LLaMA-1B 
& 13.132 \pm 0.362 
& 9.735 \pm 0.183 \\

Gemma-1B 
& 6419.709 \pm 738.524 
& 1600.649 \pm 148.048 \\

DeepSeek-1.5B 
& 137.372 \pm 6.277 
& 67.798 \pm 2.762 \\

Qwen-1.7B 
& 567.026 \pm 41.272 
& 303.196 \pm 31.500 \\

\bottomrule
\end{tabular}
\end{table}

\FloatBarrier

\section{Discussion}



\begin{figure}
    \centering
    \includegraphics[width=1\linewidth]{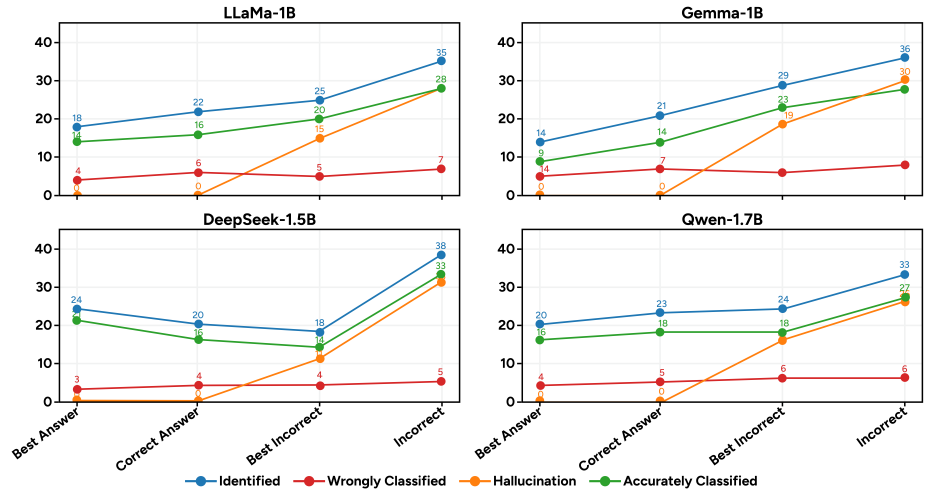}
    \caption{Component Trends Across Outcome Types (Per Model)}
    \label{fig:TREND}
\end{figure}

The figure \ref{fig:TREND} presents component trends across four small-scale language models (LLaMA-1B, Gemma-1B, DeepSeek-1.5B, and Qwen-1.7B) under four outcome types: Best Answer, Correct Answer, Best Incorrect, and Incorrect. Performance is evaluated using four metrics: Identified, Wrongly Classified, Hallucination, and Accurately Classified. Across all models, the number of accurately classified and identified instances increases from correct to incorrect categories, while hallucination rises sharply in the Best Incorrect and Incorrect outcomes. Wrong classifications remain relatively low and stable compared to hallucination. DeepSeek-1.5B and Gemma-1B show stronger gains in identification and accurate classification in the incorrect categories, whereas LLaMA-1B and Qwen-1.7B demonstrate more gradual trends. The pattern suggests that hallucination is more strongly triggered in error-prone responses, while accurate classification correlates with stronger answer quality. Importantly, all classifications (identified, accurately classified, wrongly classified, and hallucinated instances) were manually verified and cross-checked with the researchers to ensure consistency, validity, and reliability of the evaluation.

\begin{figure}
    \centering
    \includegraphics[width=0.7\linewidth]{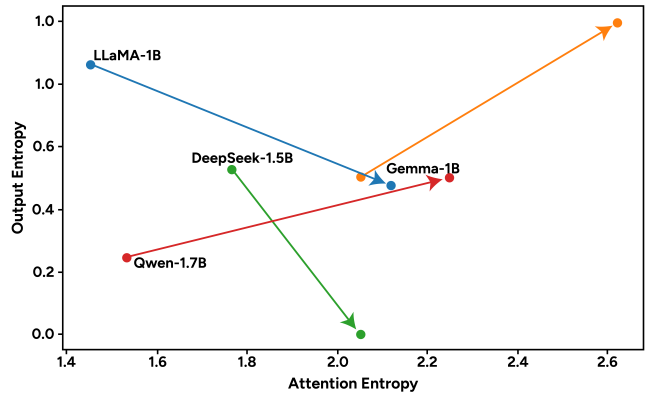}
    \caption{Entrophy Drift}
    \label{fig:entrophy}
\end{figure}

\FloatBarrier
\vspace{-0.3CM}
Figure \ref{fig:entrophy} shows that entropy evolves during generation. Attention entropy increases for all models, meaning attention becomes more diffuse toward the end of responses. However, output entropy does not follow the same pattern. DeepSeek-1.5B and LLaMA-1B reduce output entropy over time, indicating stronger confidence consolidation. In contrast, Gemma-1B and Qwen-1.7B increase output entropy in later stages, indicating growing uncertainty. Attention diffusion and prediction confidence are related but not identical processes. Models can broaden attention while simultaneously sharpening or weakening token-level certainty.
\FloatBarrier

\begin{figure}
    \centering
    \includegraphics[width=1\linewidth]{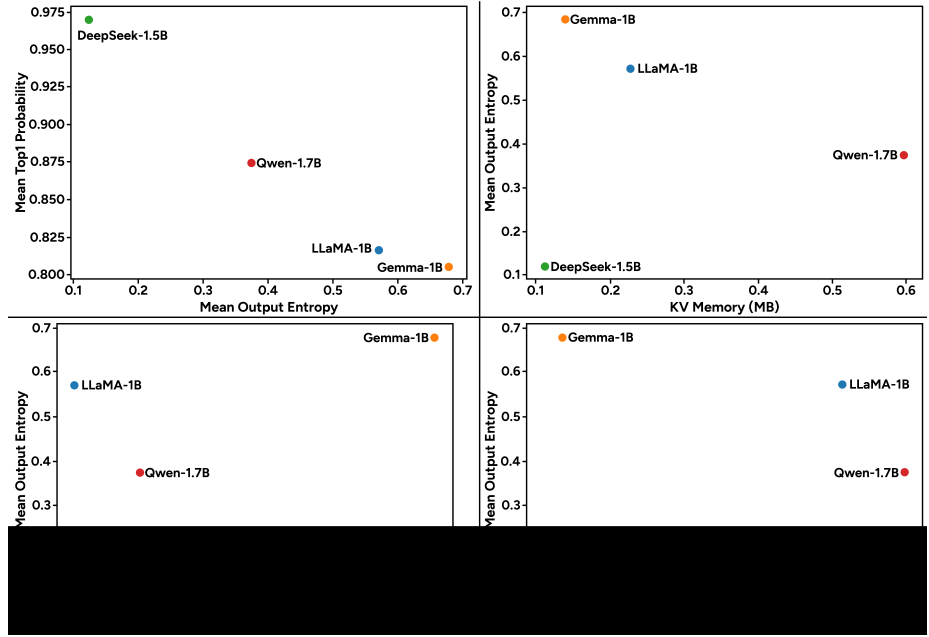}
    \caption{Relationships Between Output Entropy, Top-1 Confidence, KV Memory, Representational Drift, and Attention Dispersion Across Small Language Models}
    \label{fig:collusion}
\end{figure}

Figure \ref{fig:collusion} (Left top image) highlights a structural trade-off between entropy and Top1 probability. DeepSeek operates in a low-entropy, high-confidence regime. Gemma operates in a high-entropy, lower-confidence regime. LLaMA and Qwen fall between these extremes. When linked to Table \ref{tab:transposed_model_entropy}, these regimes align with behavioural outcomes. DeepSeek achieves the highest proportion of accurately classified Best Answers (21\%), while Gemma shows the highest hallucination proportions in incorrect categories. LLaMA and Qwen maintain more balanced profiles. Extreme determinism and extreme exploration both carry risks; calibrated confidence appears more beneficial than either extreme.

Figure \ref{fig:collusion} (right top image) shows that KV memory footprint does not directly determine entropy behavior. Qwen uses the highest memory but does not achieve the lowest uncertainty. DeepSeek uses minimal memory yet maintains strong decisiveness. Architectural design influences uncertainty more than raw memory usage. Structural efficiency and probabilistic control operate in partially independent dimensions.

\FloatBarrier

Figure \ref{fig:collusion} (left bottom image) reveals a clear relationship between hidden-state drift (Delta L2) and output entropy. Gemma exhibits the largest representational drift and the highest entropy. DeepSeek shows the smallest drift and the lowest entropy. LLaMA and Qwen lie between these extremes. Stronger geometric transformations across decoding steps are associated with higher uncertainty. However, drift magnitude is not the same as entropy. Drift reflects geometric change in hidden space, while entropy reflects probabilistic dispersion over tokens. Geometric energy and probabilistic uncertainty are coupled but conceptually distinct.

Figure \ref{fig:collusion} (right bottom image) shows that attention head specialization varies across architectures. Qwen has the highest Head Dispersion Index (HDI), indicating strong head heterogeneity. DeepSeek and Gemma show more uniform head behavior. Combined with earlier layer-level analysis, this confirms structural specialization across depth. Some layers act as broad context integrators, while others perform refinement and consolidation. Head diversity and entropy are related but not perfectly aligned, indicating that architectural specialization shapes reasoning pathways.

\begin{figure}
    \centering
    \includegraphics[width=0.5\linewidth]{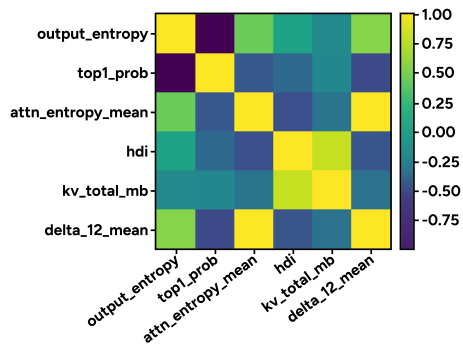}
    \caption{Correlation heatmap of entropy, confidence, attention dispersion, representation drift, and memory metrics.}
    \label{fig:figure6}
\end{figure}

Figure \ref{fig:figure6} integrates all structural measures. Output entropy shows a strong negative correlation with Top1 probability, confirming the confidence–uncertainty relationship. Entropy positively correlates with representation drift, supporting the link between geometric instability and probabilistic uncertainty. Attention entropy shows moderate association with output entropy, while KV memory shows weak direct correlation. No single structural metric explains truthfulness on its own; truthfulness emerges from interacting structural components.

\FloatBarrier

Table \ref{tab:transposed_model_entropy} connects these structural findings to behavioral outcomes. DeepSeek represents a deterministic regime with low entropy, stable geometry, and strong confidence consolidation. Gemma represents an exploratory regime with high entropy, large drift, and higher hallucination rates. LLaMA and Qwen represent balanced regimes with moderate entropy and moderate stability. Truthfulness is therefore not driven by model size alone but by how uncertainty evolves, how representations shift, and how attention is distributed during decoding.

The high inter-rater agreement strengthens these conclusions. With $\kappa = 0.81$ and $\alpha = 0.79$, the outcome classifications are reliable and not driven by subjective bias. Disagreements mainly occurred in borderline distinctions between Best and Correct answers, while severe hallucinations showed near-perfect agreement. This confirms that the structural–behavioral relationships observed are robust and not annotation artifacts.

Overall, the results demonstrate that truthfulness is a structural phenomenon emerging from the interaction between entropy evolution, attention diffusion, head specialization, and representational stability. Accuracy alone cannot capture this complexity. Generative reliability depends on how probabilistic uncertainty and geometric dynamics are internally regulated over time.

\FloatBarrier

\section{Conclusion}

This study introduced a trace-level structural framework to analyze generative reliability in SLMs. Instead of evaluating only final output accuracy, internal decoding dynamics across token-level entropy, attention diffusion, head dispersion, hidden-state magnitude, and representational drift were analyzed. By extracting structural signals at each decoding step and layer, we moved beyond outcome-level evaluation to provide insight into how uncertainty evolves during generation.

Our analysis reveals that entropy regulation is not static. Attention diffusion increased across all models during decoding, yet prediction confidence evolved differently across architectures. Some models showed decreased output entropy over time, indicating consolidation toward deterministic prediction, while others exhibited increasing entropy, reflecting exploratory behavior. Hidden-state magnitude and layer-to-layer representational drift further demonstrated that geometric transformation and probabilistic uncertainty operate as partially independent dimensions. These findings suggest that truthfulness is not determined primarily by confidence levels but emerges from dynamic interactions between entropy regulation, attention allocation, and representational geometry.

The integration of probabilistic and geometric analysis contributes a structural methodology for evaluating generative stability in SLMs. By distinguishing deterministic, exploratory, and balanced structural patterns, this study provides a principled framework for understanding architectural differences in SLMs. For future work. First, the study uses greedy decoding; extending to temperature or nucleus sampling would test how structural entropy interacts with sampling variability. Second, moving beyond TruthfulQA to multi-hop and long-form reasoning would examine behavior under deeper cognitive load. Third, analyzing step-to-step representational drift across tokens not only layer-wise drift could expose temporal instability in reasoning chains. Fourth, scaling to larger models would clarify how size affects entropy regulation and geometric stability. Finally, structural metrics can be embedded into adaptive, entropy-aware decoding and attention-regularized training, enabling uncertainty-guided optimization for building more reliable SLMs by directly controlling internal dynamics rather than only improving final accuracy.

By shifting the focus from output-only evaluation to trace-level structural dynamics, this work advances a more interpretable and principled understanding of generative reliability in SLMs.

\section{Declaration on the Use of Generative AI}
Language editing and grammar-checking tools were used to improve clarity and readability of the manuscript.

\section*{Appendix}
\section{Algorithm Description of Trace-Level Extraction}

\subsection{Prompt Standardization and Input Construction}

Instruction-tuned models require consistent formatting for predictable behavior. Some models use chat-style templates, while others accept raw text input. To ensure consistency, the tokenizer was inspected for the presence of a chat template. If available, the prompt was wrapped using the tokenizer’s \texttt{apply\_chat\_template} function. Otherwise, the prompt was tokenized directly. This normalization ensures models receive input in the format used during training. Without such standardization, formatting differences could artificially influence entropy and attention behavior. The procedure therefore isolates architectural differences rather than prompt-formatting effects.

\subsection{Decoding Protocol and Termination Strategy}

Generation was performed step-by-step using autoregressive decoding with key--value caching. After processing the full prompt, cached key and value tensors were stored for each transformer layer. During generation, only the most recently generated token was passed back into the model along with cached states, preserving full historical context while reducing computational overhead. Decoding terminated under three conditions:

\begin{enumerate}
    \item If a model-specific stop token was generated. Stop tokens were constructed by combining tokenizer-defined EOS tokens, configuration identifiers, generation configuration tokens, and common special tokens (e.g., \texttt{<eos>}, \texttt{<endoftext>}).
    \item If a maximum of 1000 tokens was reached.
    \item If the same token was generated for 15 consecutive steps, preventing degenerate repetition.
\end{enumerate}

These termination rules ensure fairness across architectures and prevent pathological generation behavior.

\subsection{Trace-Level Structural Signal Extraction}

Instead of evaluating only the final generated output, internal signals were recorded at every decoding step and for every transformer layer. Extraction occurred in two phases:

\textbf{Prompt Phase (Step 0):} A full forward pass over the input prompt was executed. Hidden states, attention tensors, logits, and key--value cache contents were extracted.

\textbf{Generation Phase:} The same signals were extracted at each decoding step. Each step produced hidden states corresponding to the embedding output and each transformer layer, along with attention tensors when enabled. 
Each decoding step generated multiple rows in the trace file—one per layer plus the embedding output—forming a long-format dataset indexed by phase, step, and layer.

\subsection{Probabilistic Uncertainty Measurement}

At each decoding step, logits were converted into probabilities using the softmax function. Token-level Shannon entropy was computed as:

\begin{equation}
H = - \sum_{i} p_i \log p_i
\end{equation}

Low entropy indicates high confidence, while high entropy reflects uncertainty. Two additional confidence measures were recorded:

\begin{itemize}
    \item Top-1 probability
    \item Top-1–Top-2 probability gap
\end{itemize}

The Top-1–Top-2 gap measures decisiveness. A larger gap indicates stronger preference for a single token.

\subsection{Attention Entropy and Diffusion}

For each transformer layer, attention weights corresponding to the last query position were extracted and normalized. Entropy was computed across attended key positions for each attention head. The mean entropy across heads was recorded as layer-level attention entropy. Lower attention entropy indicates concentrated attention, while higher entropy indicates diffusion.

\subsection{Hidden-State Magnitude and Representation Drift}

For each layer, the L2 norm of the last-token hidden state was computed. Representation drift between consecutive layers was measured as:

\begin{equation}
\Delta_l = \left\| h_l - h_{l-1} \right\|_2
\end{equation}

Larger values indicate stronger representational transformation. Entropy captures probabilistic uncertainty, whereas representation drift captures geometric movement in vector space. These dimensions are related but distinct.

\subsection{Key-Value Cache Memory Measurement}

Memory usage was computed by measuring the byte size of key and value tensors stored in the cache for each layer. Both per-layer and total memory usage were recorded. 
This enables structural scaling differences to be observed as sequence length increases and provides insight into architectural efficiency beyond entropy analysis.
%
%

\end{document}